\documentclass[10pt,twocolumn,letterpaper]{article}

\usepackage[preprint]{cvpr}      

\usepackage{graphicx}
\usepackage{amssymb}
\usepackage{pifont}

\usepackage{amsmath}
\usepackage{amssymb}
\DeclareMathOperator{\E}{\mathbb{E}}

\definecolor{cvprblue}{rgb}{0.21,0.49,0.74}
\usepackage[pagebackref,breaklinks,colorlinks,allcolors=cvprblue]{hyperref}

\def\paperID{} 
\def\confName{CVPR}
\def\confYear{2026}

\title{PhyPrompt: RL-based Prompt Refinement for Physically Plausible Text-to-Video Generation}

\author{
Shang Wu$^{1}$ \quad Chenwei Xu$^{1}$ \quad Zhuofan Xia$^{1}$ \quad Weijian Li$^{1}$ \\
Lie Lu$^{2}$ \quad Pranav Maneriker$^{2}$ \quad Fan Du$^{2}$ \quad Manling Li$^{1}$ \quad Han Liu$^{1}$\\
{\small $^{1}$ Northwestern University \quad $^{2}$ Dolby Laboratories}
}

\begin{document}
\maketitle
\begin{abstract}
State-of-the-art text-to-video (T2V) generators frequently violate physical laws despite high visual quality. We show this stems from insufficient physical constraints in prompts rather than model limitations: manually adding physics details reliably produces physically plausible videos, but requires expertise and does not scale.
We present PhyPrompt, a two-stage reinforcement learning framework that automatically refines prompts for physically realistic generation. First, we fine-tune a large language model on a physics-focused Chain-of-Thought dataset to integrate principles like object motion and force interactions while preserving user intent. Second, we apply Group Relative Policy Optimization with a dynamic reward curriculum that initially prioritizes semantic fidelity, then progressively shifts toward physical commonsense. This curriculum achieves synergistic optimization: PhyPrompt-7B reaches 40.8\% joint success on VideoPhy2 (8.6pp gain), improving physical commonsense by 11pp (55.8\% to 66.8\%) while simultaneously increasing semantic adherence by 4.4pp (43.4\% to 47.8\%). Remarkably, our curriculum exceeds single-objective training on both metrics, demonstrating compositional prompt discovery beyond conventional multi-objective trade-offs.
PhyPrompt outperforms GPT-4o (+3.8\% joint) and DeepSeek-V3 (+2.2\%, 100$\times$ larger) using only 7B parameters. The approach transfers zero-shot across diverse T2V architectures (Lavie, VideoCrafter2, CogVideoX-5B) with up to 16.8\% improvement, establishing that domain-specialized reinforcement learning with compositional curricula surpasses general-purpose scaling for physics-aware generation.
\end{abstract}    
\section{Introduction}
\label{sec:intro}
Text‑to‑video (T2V) generation has advanced rapidly, with large‑scale
models now producing visually impressive clips from short textual
prompts \cite{zheng2024open, yang2024cogvideox, chen2024videocrafter2, wang2025wan}.  
A critical shortcoming remains unresolved: state‑of‑the‑art
T2V systems frequently violate basic physical commonsense even when the
visual quality is high.  
Objects may teleport, ignore gravity, or pass
through each other, violating real‑world physics.  
Recent evaluations \cite{meng2024towards, bansal2024videophy, Bansal2025} report major physical inconsistencies, showing that top
public models still struggle with scenarios requiring mass and momentum conservation laws.  
This gap between visual realism and physical plausibility severely limits their deployment in domains that demand fidelity to real‑world dynamics (e.g.\ robotics and simulation).  

\begin{figure*}[t]
    \centering
    \includegraphics[width=\textwidth]{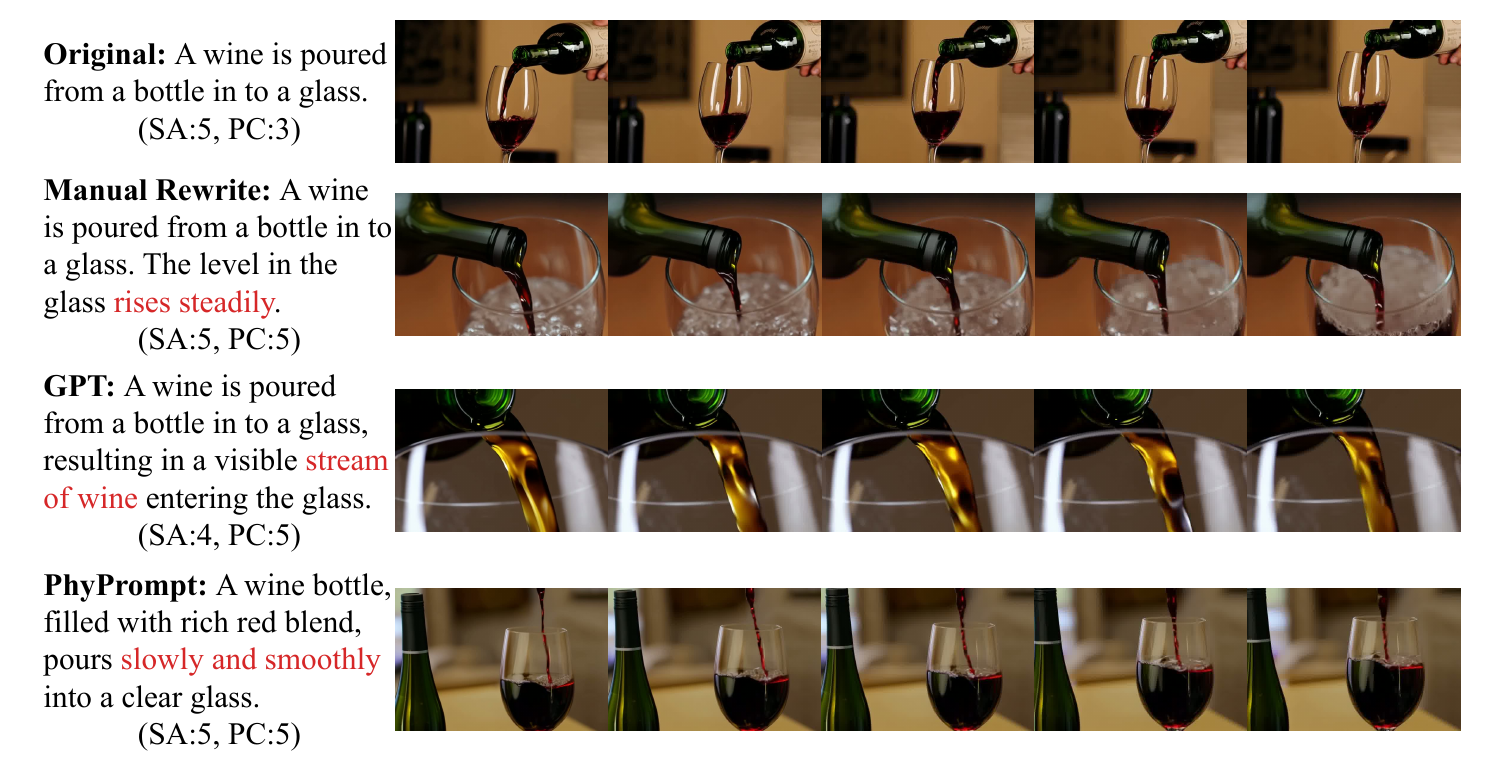}
    \caption{\textbf{Video Generated by CogVideoX-5B Using Different Prompts.} We compare the original prompt \textit{A wine is poured from a bottle in to a glass} with three enhanced versions: manual rewrite, GPT-4o, and PhyPrompt.  Semantic Adherence (SA) and Physical Commonsense (PC) scores are shown for each. The original prompt fails to depict rising liquid levels (PC:3). Manual rewriting with "rises steadily" achieves perfect scores (SA:5, PC:5). GPT-4o emphasizes the "visible stream of wine" but slightly reduces semantic alignment (SA:4, PC:5). PhyPrompt automatically generates "slowly and smoothly," matching manual rewrite quality (SA:5, PC:5) without human intervention.}
    \label{fig:demo}
\end{figure*}

One root cause is a mismatch between how T2V models are trained and how they are used. During training, models ingest detailed, carefully crafted captions paired with videos. However, real prompts are often brief or underspecified at inference, yielding suboptimal outputs \cite{xue2024phyt2v, cheng2025vpo}.
For instance, a user might write "\textit{A wine is poured from a bottle in to a glass}" and obtain a video where wine flows from the bottle but the liquid level in the glass remains nearly unchanged (Figure~\ref{fig:demo}, top row). While this prompt achieves high semantic adherence (Semantic score:5), it suffers from poor physical commonsense (Phyiscal score:3). Crucially, \emph{manually rewriting} the prompt to include explicit physical details such as "\textit{The level in the glass rises steadily}" successfully produces physically plausible videos where the liquid accumulates visibly (Figure~\ref{fig:demo}, second row), achieving perfect scores (Semantic score:5, Phyiscal score:3). This demonstrates that current T2V generators \emph{possess the capability} to synthesize physically realistic content when provided with physics-aware prompts; the bottleneck lies in the prompt itself. However, manual prompt engineering requires domain expertise, is time-consuming, and does not scale to diverse scenarios.

Existing automated approaches partially address this gap but face key limitations.
Promptist \cite{hao2023promptist} automatically enhances user
inputs for text‑to‑image models via supervised fine‑tuning followed by reinforcement learning
(RL) on aesthetic rewards, but does not target physical plausibility.
Early text‑to‑video work employs large language models (LLMs) such as GPT‑4o to enhance prompts through in‑context learning; while these methods can improve physics adherence (e.g., GPT‑4o achieves PC:5 in ~\cref{fig:demo}), they may reduce semantic fidelity (SA:4) and lack systematic optimization for video‑level physical commonsense.
VPO~\cite{cheng2025vpo} refines prompts with preference‑based RL to ensure they are harmless and accurate, but, like Promptist, does not explicitly address physical plausibility.
Very recently, PhyT2V introduced an LLM-guided, iterative self-refinement loop that enables chain-of-thought and "step-back" reasoning in T2V prompting, improving adherence to real-world physics and outperforming prior prompt enhancers on out-of-distribution scenarios~\cite{xue2024phyt2v}.  
While PhyT2V demonstrates the value of iterative LLM feedback, it requires multiple prompting rounds and a complex step-back mechanism, limiting its efficiency.

To address these limitations, we introduce \textbf{PhyPrompt}, a novel framework that leverages a Large Language Model (LLM) trained with reinforcement learning (RL) to \emph{automatically} transform user prompts into descriptions that elicit physically realistic video outputs, matching the quality of manual physics-aware prompt engineering without requiring human expertise. 
Our approach is characterized by a two-stage training pipeline. 
First, the LLM undergoes Supervised Fine-Tuning (SFT) on a high-quality, Chain-of-Thought (CoT) dataset curated by us, based on PhyGenBench \cite{meng2024towards}.
This dataset is specifically designed to imbue the LLM with the ability to reason about physical phenomena and translate these considerations into descriptive text suitable for video generation. 
In the second stage, the SFT-trained LLM's prompt generation policy is further refined using Group Relative Policy Optimization (GRPO)~\cite{guo2025deepseek}, an RL algorithm that efficiently explores the prompt space by sampling multiple candidates per query without requiring a separate value network.

A key innovation in PhyPrompt is its dynamic reward mechanism inspired by curriculum training \cite{bengio2009curriculum}. 
The GRPO training is guided by a composite reward signal that adaptively balances the physical commonsense of the generated video and its semantic alignment with the user's intent.
In our design, the reward prioritizes semantic alignment in the early stages of training, gradually shifting the emphasis towards physical commonsense as the LLM's policy matures.
This curriculum-like approach ensures that the LLM first learns to preserve the user's intent before focusing on the more nuanced task of enhancing physical realism.
Our architecture leaves the video generator frozen and trains only a lightweight, model‑agnostic rewriter that converts user prompts into physics‑aware descriptions.
This avoids costly generator fine‑tuning and allows one rewriter to serve multiple back‑ends.  

We validate our approach on four cutting‑edge generators: Lavie~\cite{wang2023lavie}, VideoCrafter2~\cite{chen2024videocrafter2}, CogVideoX-2B and CogVideoX-5B~\cite{wang2024cogvideodex}.
For each, we insert the trained LLM between the user and the generator.  
Experiments on the test dataset from VideoPhy2 dataset show that rewritten prompts reduce physics violations while maintaining visual quality and semantic fidelity.
As demonstrated in~\cref{fig:demo}, PhyPrompt automatically generates prompts that achieve the same level of physical plausibility (PC:5) and semantic adherence (SA:5) as manually engineered physics-aware prompts, eliminating the need for human expertise in prompt design.
The main contributions of this work are as follows:
\begin{itemize}
    \item We demonstrate that current T2V generators can produce physically plausible videos when provided with physics-aware prompts, and propose PhyPrompt, a novel two-stage trained LLM system.
    \item We introduce a dynamic, time-dependent reward mechanism for RL-based prompt optimization that progressively shifts focus from semantic alignment to physical commonsense, facilitating a balanced and effective learning curriculum.
      \item Empirical validation showing zero-shot transfer across multiple T2V models, with joint semantic-adherence and physical-commonsense gains of up to 16.8\%, achieving quality comparable to manual physics-aware prompt engineering.
\end{itemize}
\section{Related Work}
\label{sec:related}

\noindent\textbf{Reinforcement Learning for LLM Reasoning.}  
RL now complements supervised fine-tuning (SFT) to enhance LLM reasoning.  
Instability and cost in PPO have motivated alternatives such as Group Relative Policy Optimization (GRPO)~\cite{shao2024grpo,li2025mogrp} and Direct Preference Optimization (DPO)~\cite{goffer2023dpo}.  
Typical pipelines start from an SFT checkpoint with Chain-of-Thought traces and apply GRPO (or derivatives) to improve factual consistency and deductive depth~\cite{yue2025cot,zhang2025sgrpo}, often incorporating RL from Human Feedback (RLHF)~\cite{outrun2023rlhf} or Reinforcement Learning from AI Feedback (RLAIF)~\cite{rajesh2024rlaif}.  
Multi-objective settings introduce label-specific rewards or curricula that adapt weights during training~\cite{chen2025emorl,shibata2025dynamicrl}, and recent work has explored hierarchical reward decomposition to scale reasoning depth~\cite{nakano2023hapt}.  
PhyPrompt follows this two-stage recipe: SFT on a physics-focused CoT corpus, then GRPO with a curriculum reward that gradually shifts emphasis from semantic alignment to physical-commonsense cues vital for video generation.

\noindent\textbf{Benchmarks for Physical Commonsense in Video Generation.}  
To judge physics realism, purpose-built datasets replace pixel-level metrics.  
VideoPhy2 provides human-plus-automatic scores for Semantic Adherence (SA) and Physical Commonsense (PC)~\cite{bansal2025videophy2}, while PhyGenBench widens coverage to 160 prompts across 27 physical laws~\cite{phygenbench2024}.  
PhyCoBench tracks coherence in seven principle categories via optical-flow signals~\cite{chen2025phycobench}, and IPV-Bench stresses models with intentionally impossible scenes~\cite{wang2025ipvbench}.  
Other recent benchmarks include VisualPhysicist for dynamic object interactions~\cite{singh2024visualphysicist} and SpeedNet for temporal consistency in motion~\cite{kim2023speednet}.  
These resources form the first quantitative yardsticks for physics-aware video synthesis.

\noindent\textbf{Prompt Rewriting for Image and Video Generation.}  
Promptist uses RL-based adaptation to refine text-to-image prompts~\cite{hao2023promptist}, and DiffusionPrompt employs gradient-based style rewrites~\cite{lee2023diffusionprompt}. In text-to-video, VPO combines SFT and DPO with multi-level feedback~\cite{cheng2025vpo}, Prompt-A-Video applies reward-driven caption evolution~\cite{ji2024prompt}, MotionPrompt adds optical-flow temporal guidance~\cite{nam2024motionprompt}, MotionCraft and ReVision inject physics cues at generation~\cite{agnolucci2024motioncraft,wang2025revision}, and newer approaches, VideoPromptTuning~\cite{xu2024vpt}, Prompt2Video~\cite{fernandez2024prompt2video}, and PhyT2V~\cite{xue2024phyt2v}, use lightweight adapters or iterative CoT reasoning for enhanced physical adherence.  

PhyPrompt differs by pre-training on a physics-focused CoT corpus and maximising an automatic PC reward computed on the generated video, yielding a lightweight, generator-agnostic front end for physics-aware T2V models.  
Besides, PhyPrompt is the first prompt rewriter that trains end-to-end with RL on video-based physical-commonsense rewards and transfers zero-shot across multiple heterogeneous T2V generators without re-tuning.  
These qualities position PhyPrompt as a practical bridge between user intent and physics-consistent video generation.

\section{Method}
\label{sec:method}

\begin{figure*}
    \centering
   \resizebox{1.0\textwidth}{!}{%
    \includegraphics{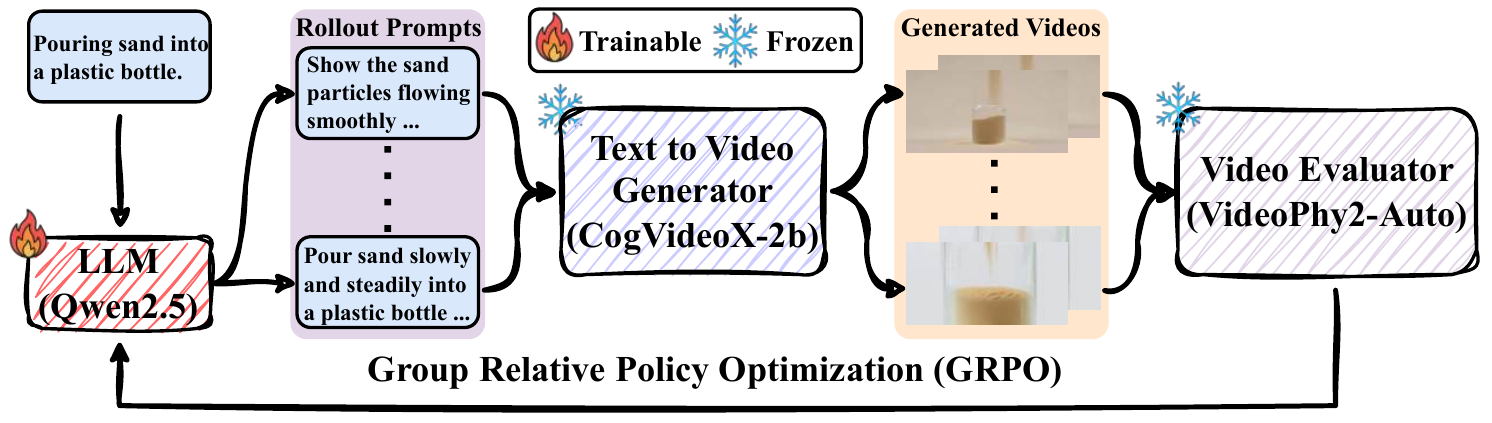}
    }
    \vspace{-0.1in}
    \caption{\textbf{PhyPrompt GRPO Pipeline.} For each input prompt, the LLM (Qwen2.5~\cite{Qwen2024}) generates multiple enhanced prompts and for each enhanced prompt, T2V generator (CogVideoX-2B~\cite{yang2024cogvideox}) generates one video. Video evaluator (VideoPhy2-Auto~\cite{bansal2025videophy2}) scores each video. We utilize GRPO to optimize the LLM.}
    \label{fig:pipeline}
    \vspace{-0.2in}
\end{figure*}

This section describes (i) our physics-focused Chain-of-Thought (CoT) dataset for supervised fine-tuning, (ii) GRPO-based prompt optimization, and (iii) our dynamic reward curriculum that resolves the inherent conflict between semantic fidelity and physical realism.

\subsection{Physics-Focused Chain-of-Thought Dataset}

We construct a CoT dataset to teach language models physics-aware prompt enhancement. Starting with 160 examples from PhyGenBench \cite{phygenbench2024}, each contains: an \emph{original prompt} $x$, an \emph{physical law} $L$, and an \emph{enhanced prompt} $y_{\text{GPT}}$ from GPT-4o \cite{achiam2023gpt-4o}. Since these lack explicit reasoning, we use GPT-4o1-preview \cite{jaech2024o1-preview} to generate step-by-step reasoning chains explaining how $x$ becomes $y_{\text{GPT}}$ via $L$.

The resulting dataset contains triplets $(x, \mathbf{r}, y_{\text{CoT}})$ where $\mathbf{r}$ is the reasoning chain and $y_{\text{CoT}}$ is the physics-aware prompt. This enables supervised fine-tuning of Qwen2.5 while preserving user intent.

\subsection{Two-Stage Training Pipeline}

\noindent\textbf{Supervised Fine-Tuning.} We initialize $\pi_{\theta}$ by minimizing cross-entropy loss:
\begin{equation}
  \mathcal{L}_{\text{SFT}}
  = -\E_{(x,y_{\text{CoT}})} \bigl[\log \pi_{\theta}(y_{\text{CoT}} \mid x)\bigr].
\end{equation}

\noindent\textbf{Reinforcement Learning via GRPO.} Figure~\ref{fig:pipeline} shows our pipeline. User prompt $x$ is rewritten by $\pi_{\theta}$ into enhanced prompt $y$, fed to a fixed T2V generator $G$ (CogVideoX-2B \cite{yang2024cogvideox}). The video $v = G(y)$ is scored by VideoPhy2-AutoEval \cite{bansal2025videophy2}, driving RL to refine $\pi_{\theta}$ without altering $G$.
For each prompt $x^{(j)}$ in batch size $B$, we sample $G=4$ candidate rewrites from $\pi_{\theta_{\text{old}}}$. Each yields video $v_i^{(j)} = G(y_i^{(j)})$ scored as $r_i^{(j)}$. Define group mean $\bar r^{(j)} = \frac{1}{G}\sum_{i=1}^G r_i^{(j)}$ and advantage:
\begin{equation}
  A_i^{(j)} = r_i^{(j)} - \bar r^{(j)}. 
  \label{eq:advantage}
\end{equation}

We update $\pi_\theta$ via clipped GRPO loss with importance ratio 
$w_i^{(j)} = \pi_{\theta}(y_i^{(j)}\mid x^{(j)})/\pi_{\theta_{\text{old}}}(y_i^{(j)}\mid x^{(j)})$:
\begin{align}
  \mathcal{L}_{\text{GRPO}}
  &= -\frac{1}{BG}\sum_{j=1}^B\sum_{i=1}^G
    \min\bigl(
      w_i^{(j)} A_i^{(j)}, \nonumber \\
    &\quad \text{clip}(w_i^{(j)}, 1-\epsilon, 1+\epsilon)\,A_i^{(j)}
    \bigr) 
    + \beta\,D_{\mathrm{KL}}[\pi_{\theta_{\text{old}}}\|\pi_{\theta}],
\end{align}
where $\epsilon=0.2$ and $\beta$ weights KL divergence to the SFT initialization.

\subsection{Dynamic Multi-Objective Reward Design}

Semantic adherence (SA) and physical commonsense (PC) are inherently conflicting: single-objective optimization exhibits severe negative transfer where optimizing SA degrades PC and vice versa (Section~\ref{sec:reward_design}). Static weighted combinations cannot jointly maximize both objectives. We propose a dynamic reward curriculum that achieves synergistic optimization, exceeding the upper bounds of either single-objective approach.

\noindent\textbf{Reward Components.}
VideoPhy2-AutoEval \cite{bansal2025videophy2} provides two scores on a 1-5 Likert scale: \textbf{Semantic Adherence} ($r_{\text{sa}}$) measures video-prompt alignment, and \textbf{Physical Commonsense} ($r_{\text{pc}}$) measures physical plausibility.

\noindent\textbf{Dynamic Curriculum.}
The composite reward uses time-dependent weights:
\begin{equation}
  R(t) = w_{\text{sa}}(t) \cdot r_{\text{sa}} + w_{\text{pc}}(t) \cdot r_{\text{pc}},
  \label{eq:total_reward}
\end{equation}
with exponential decay:
\begin{equation}
  w_{\text{sa}}(t)=\exp\bigl(-\alpha\,t/T\bigr),\quad
  w_{\text{pc}}(t)=1-w_{\text{sa}}(t),
  \label{eq:weights}
\end{equation}
where $t$ is the training step, $T$ is total steps, and $\alpha$ controls decay rate.

\noindent\textbf{Compositional Learning Mechanism.}
Early training ($w_{\text{sa}} \approx 1$) establishes semantic scaffolding: object identities, relationships, and scene structure. Late training ($w_{\text{pc}} \approx 1$) refines this scaffolding with physical specificity: forces, dynamics, and causal interactions. This staged approach discovers prompt structures unreachable by single-objective optimization~\cite{bengio2009curriculum}. 

Semantic grounding provides compositional scaffolding~\cite{scholkopf2021toward}, while physical constraints act as implicit regularization~\cite{liebel2018auxiliary}. SA-only optimization lacks refinement pressure from physics; PC-only optimization lacks semantic coherence to organize physical details. Our curriculum leverages both sequentially, achieving superadditive performance that simultaneously exceeds both single-objective upper bounds (in Section~\ref{sec:reward_design}). This demonstrates that our method discovers novel compositional prompt structures beyond static Pareto frontiers, aligning with recent multi-stage RL paradigms~\cite{guo2025deepseek} where progressive task decomposition unlocks capabilities beyond direct optimization.
\section{Experiments}

This section details the experimental methodology employed to evaluate the effectiveness of PhyPrompt on physics-driven video generation tasks. Our assessment focuses on three primary aspects: (1) the physical plausibility and semantic fidelity of the generated videos, (2) the transferability of the learned PhyPrompt rewriter across diverse text-to-video (T2V) generators, and (3) ablation studies investigating the contributions of key components within our proposed approach.
\\
\textbf{Text-to-Video Generators.}
PhyPrompt's performance is evaluated using several state-of-the-art T2V diffusion models, with parameter counts ranging from 860 million to 5 billion:

\begin{itemize}
\item \textbf{Lavie}~\cite{wang2023lavie}: An 860M-parameter model that utilizes cascaded latent diffusion, trained on the Vimeo25M dataset. It is capable of generating videos up to 8 seconds in length at a 1280×720 pixel resolution.
\item \textbf{VideoCrafter2}~\cite{chen2024videocrafter2}: This 1.2B-parameter model employs a diffusion-based architecture. It was trained using a combination of low-quality videos and high-quality images to improve motion consistency and visual fidelity, producing videos up to 8 seconds in duration at a 512×512 resolution.
\item \textbf{CogVideoX-2B}~\cite{yang2024cogvideox}: A 2B-parameter model featuring a 3D variational autoencoder and an expert transformer with adaptive LayerNorm. It generates videos up to 6 seconds long at a 720×480 resolution.
\item \textbf{CogVideoX-5B}~\cite{yang2024cogvideox}: This 5B-parameter model shares the architecture of CogVideoX-2B but benefits from larger-scale training data. It produces videos up to 6 seconds in length at a 720×480 resolution.
\end{itemize}
To ensure consistent evaluation across all models, generated videos were standardized to a duration of 6 seconds, a frame rate of 4 FPS, and a resolution of 720×480 pixels. Detailed configurations of the T2V generation setup are provided in supplementary material.
\\
\textbf{Prompt Enhancer.}
PhyPrompt variants are obtained by fine‑tuning Qwen2.5‑Instruct checkpoints with 1.5B, 3B, and 7B parameters.  
For comparison we include four automatic prompt‑refinement baselines: Promptist~\cite{hao2023promptist}, PhyT2V~\cite{xue2024phyt2v}, GPT‑4o~\cite{achiam2023gpt-4o}, and DeepSeek‑V3~\cite{liu2024deepseek-v3}.
\\
\textbf{Dataset.}
Our approach is evaluated on the VideoPhy2 benchmark~\cite{bansal2025videophy2}. This benchmark comprises a diverse collection of prompts specifically designed to assess physical commonsense in video generation, encompassing scenarios related to gravity, collisions, fluid dynamics, and other fundamental physical phenomena.
\\
\textbf{Evaluation Metrics.}
Consistent with~\cite{bansal2025videophy2}, generated videos are assessed based on two criteria: semantic adherence (SA) and physical commonsense (PC). Each criterion is evaluated on a 5-point Likert scale, ranging from Very Unlikely (1) to Very Likely (5). We random sample 500 prompts from the test set and report the percentage of generated videos satisfying the following conditions:
\begin{itemize}
\item \textbf{Semantic Adherence (SA $\geq$ 4)}: The percentage of videos that accurately depict the intended content described in the prompt.
\item \textbf{Physical Commonsense (PC $\geq$ 4)}: The percentage of videos that realistically adhere to fundamental physical principles.
\item \textbf{Joint Performance (SA $\geq$  4 and PC $\geq$  4)}: The percentage of videos that concurrently achieve high scores (i.e., $\geq$ 4) in both semantic adherence and physical realism.
\end{itemize}

\subsection{Quality Assessment of Generated Videos}
\label{sec:quality}
We measure how well each prompt-rewriting strategy improves \textbf{semantic adherence (SA)} and \textbf{physical commonsense (PC)} when paired with \textbf{CogVideoX-2B}\cite{yang2024cogvideox}.  
Table\ref{tab:quality} reports results on 500 prompts from the \textsc{VideoPhy2} benchmark~\cite{bansal2025videophy2}.

\begin{table*}
\centering
\caption{\textbf{VideoPhy2 quality results with CogVideoX-2B.}  
For each prompt-rewriting strategy we report the percentage of videos scoring $\ge\!4$ for semantic adherence (SA) and physical commonsense (PC), their joint percentage (SA \& PC), and the average SA/PC scores (Avg SA, Avg PC).}
\vspace{0.1in}
\label{tab:quality}
\setlength{\tabcolsep}{3mm}{
\renewcommand\arraystretch{0.9}
\resizebox{\linewidth}{!}{
\begin{tabular}{lccccc}
\toprule
\textbf{Enhanced Model} &
\textbf{SA (\%)} $\uparrow$ &
\textbf{PC (\%)} $\uparrow$ &
\textbf{SA \& PC (\%)} $\uparrow$ &
\textbf{Avg SA} $\uparrow$ &
\textbf{Avg PC} $\uparrow$ \\
\midrule
N/A (Original Prompt)           & 43.4 & 55.8 & 32.2 & 3.842 & 4.410 \\
Promptist~\cite{hao2023promptist}     & 41.2 & 58.2 & 30.2 & 3.782 & 4.452 \\
PhyT2V~\cite{xue2024phyt2v}     & 45.0 & 62.0 & 36.2 & 3.854 & 4.522 \\			
\midrule
Qwen2.5-1.5B              & 37.6 & 64.0 & 31.8 & 3.616 & 4.562 \\
PhyPrompt-1.5B            & 42.8 & 65.4 & 37.4 & 3.800 & 4.580 \\
\addlinespace[4pt]
Qwen2.5-3B                & 40.0 & 64.2 & 32.0 & 3.754 & 4.582 \\
PhyPrompt-3B              & 46.4 & 64.8 & 39.6 & 3.868 & 4.556 \\
\addlinespace[4pt]
Qwen2.5-7B                & 44.8 & 64.6 & 34.2 & 3.848 & 4.586 \\
PhyPrompt-7B              & 47.8 & \textbf{66.8} & \textbf{40.8} & 3.878 & \textbf{4.590} \\
\midrule
GPT-4o                    & 47.0 & 60.0 & 37.0 & 3.932 & 4.488 \\
DeepSeek-V3               & \textbf{48.4} & 64.6 & 38.6 & \textbf{3.968} & 4.564 \\
\bottomrule
\end{tabular}%
}}
\end{table*}

\noindent \textbf{Effectiveness of PhyPrompt.}  
Promptist~\cite{hao2023promptist} raises PC from 55.8\% to 58.2\% but drops SA from 43.4\% to 41.2\%, causing its joint score to fall from 32.2\% to 30.2\%, confirming that a generic text-to-image adapter does not reliably balance semantics and physics in the video domain. PhyT2V~\cite{xue2024phyt2v} achieves 45.0\% SA and 62.0\% PC (36.2\% joint), demonstrating the value of iterative LLM feedback, but requires multiple prompting rounds and complex step-back mechanisms. In contrast, all three PhyPrompt checkpoints outperform both the original prompts and existing baselines on every metric. The 7B model delivers the largest gains with SA 47.8\%, PC 66.8\%, and a joint success rate of 40.8\%, representing an absolute improvement of +8.6\% over the baseline and +10.6\% over Promptist. Notably, even the 1.5B variant achieves a 37.4\% joint score, surpassing Promptist by +7.2\% while using only 1/10 the parameters of GPT-4o. The consistent improvements across all model sizes validate the effectiveness of our two-stage training approach combining physics-focused SFT with dynamic reward-based RL.

\noindent\textbf{Advantages over general-purpose large language models.}  
PhyPrompt demonstrates that domain-specialized training outperforms raw parameter scaling for physics-aware video generation. Compared to GPT-4o~\cite{achiam2023gpt-4o}, PhyPrompt-7B achieves nearly identical SA (47.8\% vs 47.0\%) while delivering substantially higher PC (66.8\% vs 60.0\%) and joint performance (40.8\% vs 37.0\%). This +6.8\% PC advantage stems from three factors: (i) end-to-end RL optimization on video quality metrics rather than text likelihood~\cite{ouyang2022training}, (ii) physics-focused CoT pretraining that encodes domain-specific reasoning~\cite{wei2022chain}, and (iii) dynamic reward curricula that balance dual objectives~\cite{bengio2009curriculum}. DeepSeek-V3~\cite{liu2024deepseek-v3}, despite 100× more parameters (671B), trails PhyPrompt-7B in PC (64.6\% vs 66.8\%) and joint performance (38.6\% vs 40.8\%), confirming that general-purpose pretraining does not substitute for task-specific feedback~\cite{hoffmann2022training}. PhyPrompt-3B matches GPT-4o using less than 1\% of its parameters, exemplifying recent findings that specialized smaller models can surpass general larger ones on focused tasks~\cite{muennighoff2023scaling}. Overall, our results align with emerging evidence that domain-targeted training with direct task feedback provides a more parameter-efficient path than scaling alone~\cite{kaplan2020scaling,wei2022emergent}.

\subsection{Zero-Shot Transfer Across Generators}
\label{sec:transfer}
\begin{figure*}[t]
    \centering
   \resizebox{1.0\textwidth}{!}{%
    \includegraphics{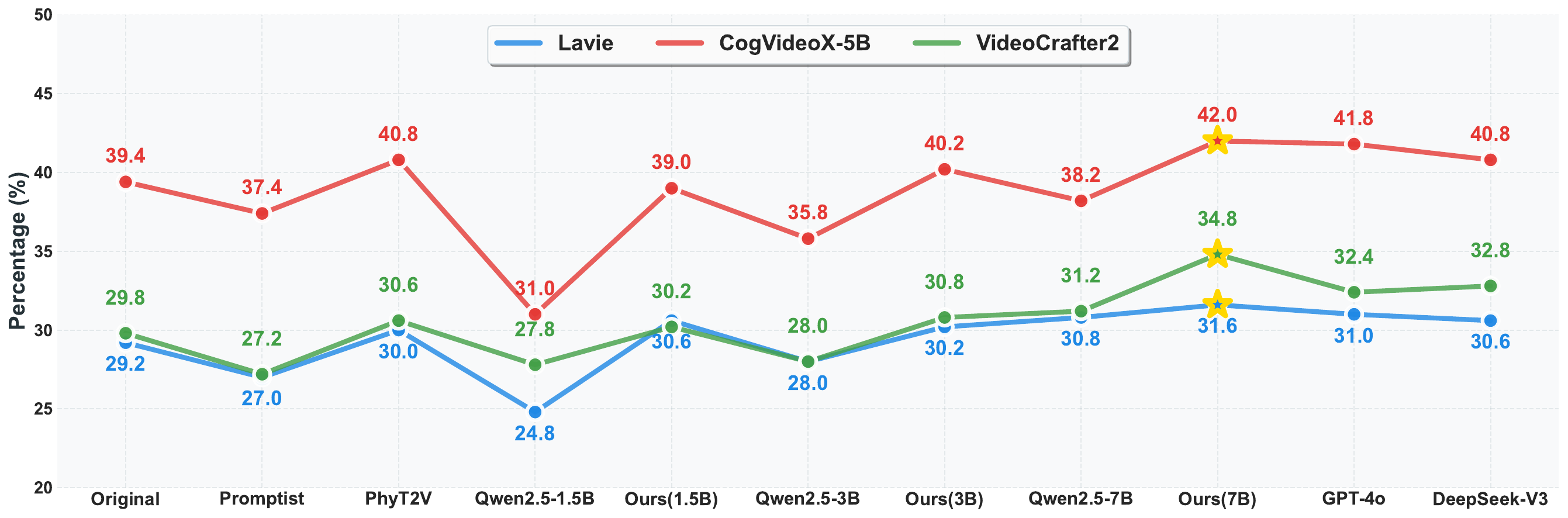}
    }
    \vspace{-0.15in}
    \caption{\textbf{Cross-Generator Transfer of PhyPrompt.} Performance comparison across different prompt enhancement methods on three text-to-video generation models (Lavie, CogVideoX-5B, and VideoCrafter2) measured by joint Physical Commonsense and Semantic Adherence (PC \& SA) metrics. Our method consistently outperforms baseline approaches (Original, Promptist, PhyT2V) and similarly-sized Qwen2.5 models across all three video generation backbends. The highest performance (42.0\%) is achieved by our 7B model on CogVideoX-5B. Gold stars indicate the best-performing method for each generator.}
    \label{fig:transfer}
    \vspace{-0.2in}
\end{figure*}

\begin{figure}[!htbp]
  \centering
    \includegraphics[width=1.0\linewidth]{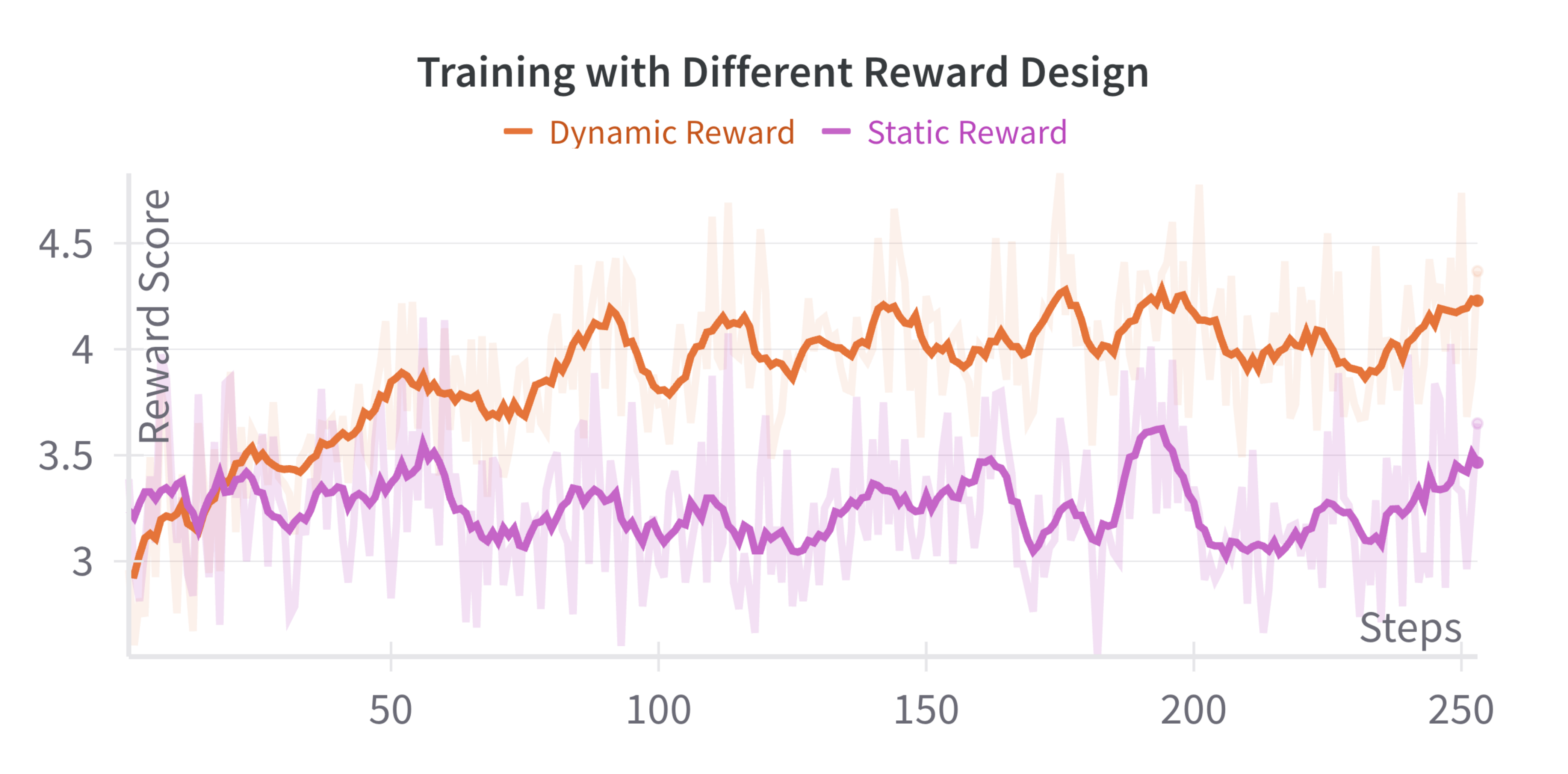}
    \captionof{figure}{Training reward curves under the dynamic (orange) and static (purple) reward formulations. Each line is smoothed with a 10-step moving average; lightly shaded traces show the unsmoothed per-episode rewards. The dynamic curriculum converges more rapidly and plateaus higher (4.2) than the static baseline (3.5), indicating more effective optimization.}
    \label{fig:reward_curve}
  \vspace{-1em}
\end{figure}

To verify PhyPrompt's generator-agnostic transfer, we evaluate PhyPrompt-7B (trained solely on CogVideoX-2B) on three architecturally diverse diffusion models: Lavie (860M), VideoCrafter2 (1.2B), and CogVideoX-5B (5B). We generate 500 videos per method for each generator and report joint SA\&PC scores in~\cref{fig:transfer}. Full results appear in the supplement.

\noindent\textbf{Joint-metric improvement.} Original prompts achieve baseline joint scores of 29.2\% (Lavie), 29.8\% (VideoCrafter2), and 39.4\% (CogVideoX-5B). Promptist degrades performance to 27.0\%, 27.2\%, and 37.4\%, respectively. PhyPrompt substantially surpasses baselines on every generator: Lavie reaches 31.6\% (+8.2\% over original), VideoCrafter2 achieves 34.8\% (+16.8\% over original), and CogVideoX-5B attains 42.0\% (+6.6\% over original). Smaller PhyPrompt variants (1.5B and 3B) exhibit the same improvement pattern, with the 3B model achieving 30.2\%, 30.8\%, and 40.2\% respectively, confirming that our training approach scales effectively across model sizes.

\noindent\textbf{Comparison with other rewriters.} PhyT2V shows improvements over Promptist but underperforms PhyPrompt across all generators, achieving 30.0\%, 30.6\%, and 40.8\% on Lavie, VideoCrafter2, and CogVideoX-5B respectively. GPT-4o, DeepSeek-V3, and size-matched Qwen baselines similarly trail PhyPrompt on the joint metric. On VideoCrafter2, the best alternative (DeepSeek-V3, 32.8\%) remains 2.0\% below PhyPrompt-7B; similar margins hold on Lavie and CogVideoX-5B.

\noindent\textbf{Zero-shot transfer.} All PhyPrompt variants are fine-tuned solely on CogVideoX-2B outputs yet yield substantial gains on Lavie and VideoCrafter2 without any back-end adjustment. This confirms genuine zero-shot transfer where physics-aware rewrites capture model-agnostic physical priors rather than exploiting generator-specific biases. Related work in parameter-efficient prompt tuning for language models has similarly demonstrated cross-model generalization~\cite{lester2021power,li2021prefix}, and RL-based prompt adapters have been shown to transfer across modalities~\cite{hao2023promptist}. By disentangling physical constraints from any particular diffusion architecture, PhyPrompt removes the need for per-generator fine-tuning, cutting computational overhead and enabling rapid deployment across diverse T2V pipelines.

\subsection{Ablation Study}
To disentangle the sources of PhyPrompt’s gains, we conduct two complementary ablations.  
First, we examine how the choice of reward schedule during GRPO fine‑tuning shapes performance.  
Second, we test the necessity of the two‑stage pipeline (SFT and RL).  
Results show that both the curriculum reward and the two‑stage procedure contribute sizable, complementary improvements.

\subsubsection{Reward Design}
\label{sec:reward_design}

\begin{figure*}[!htbp]
  \centering
  \includegraphics[width=\textwidth]{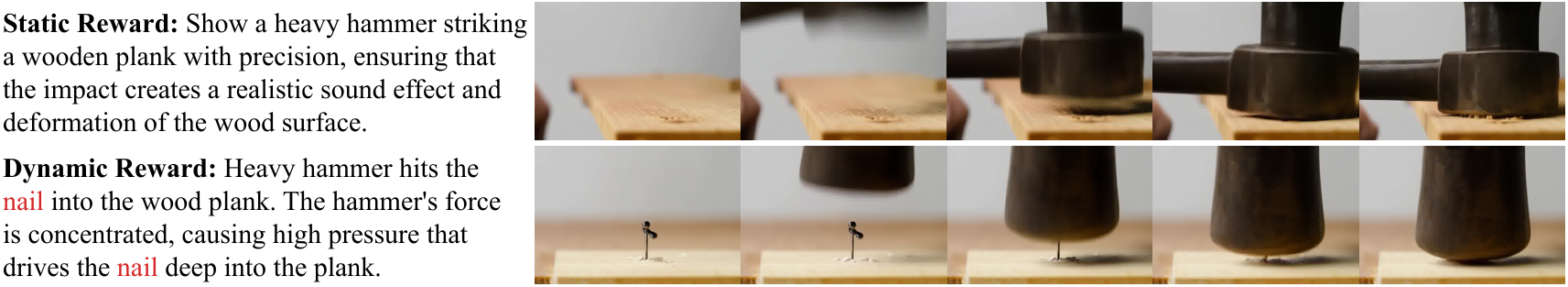}
   \caption{\textbf{Frame‐by‐frame comparison of CogVideoX-2B outputs for the “hammer and nail” scenario.}  
Top row shows results using the static‐reward prompt  
“\textit{Show a heavy hammer striking a wooden plank with precision, ensuring that the impact creates a realistic sound effect and deformation of the wood surface.}”  
The hammer impacts the plank but no nail is visible or driven.  
Bottom row shows results using the dynamic‐reward prompt  
“\textit{Heavy hammer hits the nail into the wood plank. The hammer’s force is concentrated, causing high pressure that drives the nail deep into the plank.}”   
Here the nail appears in frame 1, is pressed gradually into the wood across frames 2–3, and achieves full penetration by frame 4, with realistic wood deformation and depth of entry.}
\vspace{-1em}
  \label{fig:hammer_frames}
\end{figure*}

To demonstrate that semantic adherence (SA) and physical commonsense (PC) are inherently conflicting objectives, we conduct a systematic ablation comparing four reward strategies: (i)~SA-only optimization ($w_{\text{sa}} = 1.0, w_{\text{pc}} = 0.0$), (ii)~PC-only optimization ($w_{\text{sa}} = 0.0, w_{\text{pc}} = 1.0$), (iii)~static balanced weighting ($w_{\text{sa}} = w_{\text{pc}} = 0.5$), and (iv)~our dynamic curriculum (Eq.~\ref{eq:total_reward}). We use Qwen2.5-7B-Instruct as the base model and CogVideoX-2B as the video generator. Results are shown in \cref{fig:reward_curve} and~\cref{tab:reward_ablation}.

\noindent\textbf{Negative Transfer Between Objectives.}
\cref{tab:reward_ablation} reveals severe negative transfer. SA-only training improves semantic adherence to 45.6\% (+2.2\% over baseline) but degrades physical commonsense to 52.8\% ($-3.0\%$), yielding worse joint performance (31.6\% vs. 32.2\% baseline). Conversely, PC-only training achieves 65.2\% physics score (+9.4\%) but drops semantic adherence to 39.4\% ($-4.0\%$), with only marginal joint improvement (33.4\%). This antagonistic relationship demonstrates that SA and PC cannot be jointly maximized through naive single-objective optimization. Static balanced weighting achieves 35.4\% joint success, superior to either extreme but still 5.4\% below our dynamic curriculum.

\noindent\textbf{Dynamic Curriculum Exceeds Single-Objective Upper Bounds.}
Our dynamic curriculum achieves SA = 47.8\% and PC = 66.8\%. Remarkably, this outperforms SA-only training on semantic adherence (47.8\% vs. 45.6\%) and simultaneously outperforms PC-only training on physical commonsense (66.8\% vs. 65.2\%). In other words, our method beats each specialist on its own objective. This indicates that optimizing both objectives together with a curriculum yields better results than optimizing either objective alone~\cite{caruana1997multitask,yu2020gradient,standley2020tasks}.

We hypothesize that semantic descriptions establish object identities and relationships (e.g., "a ball" and "an incline"), while physical constraints refine these with dynamics (e.g., "accelerates" and "under gravity"). High-quality physical descriptions require precise semantic grounding~\cite{scholkopf2021toward}: one cannot specify realistic forces without first establishing what objects exist. SA-only optimization discovers prompts with good semantic structure but lacks refinement pressure from physical constraints. PC-only optimization generates physically detailed prompts but without semantic coherence to organize these details effectively. Our curriculum leverages both signals: SA establishes compositional scaffolding, then PC refines this structure with physical specificity, acting as implicit regularization~\cite{liebel2018auxiliary}. This staged approach discovers prompts in regions unreachable by single-objective trajectories~\cite{bengio2009curriculum}, explaining how we exceed both upper bounds.

Compared to static balanced weighting, our dynamic curriculum improves across all three VideoPhy2 metrics, demonstrating that the time-dependent schedule enhances both objectives rather than trading one off against the other. Notably, the curriculum achieves higher SA than even SA-only training (47.8\% vs. 45.6\%), suggesting that the staged approach enables more effective exploration of the prompt space.

\noindent\textbf{Case Study: Hammer and Nail.}
\cref{fig:hammer_frames} illustrates the qualitative impact. For the prompt "\textit{Heavy hammer hits the nail into the wood plank}," static balanced weighting produces videos where the hammer impacts the plank but the nail is absent (a semantic failure). In contrast, our dynamic curriculum generates prompts that explicitly name the nail and describe force dynamics, resulting in physically coherent sequences where the nail is progressively driven into the wood with realistic deformation. This exemplifies how our curriculum avoids the semantic drift inherent in PC-only optimization while capturing physical details that SA-only training misses.

\begin{table}[t]
  \centering
  \caption{Reward weighting ablation on \textsc{VideoPhy2}. 
           Fixed-weight strategies exhibit SA-PC trade-offs; 
           our dynamic curriculum (\cref{eq:weights}) achieves Pareto dominance.}
  \label{tab:reward_ablation}
  \vspace{0.05in}
  \small
  \begin{tabular}{lccc}
    \toprule
    \textbf{Reward Function} & \textbf{SA (\%)} & \textbf{PC (\%)} & \textbf{Joint (\%)} \\
    \midrule
    No training (baseline)     & 43.4 & 55.8 & 32.2 \\
    \midrule
    SA score              & 45.6 & 52.8 & 31.6 \\
    PC score              & 39.4 & 65.2 & 33.4 \\
    Static         & 45.0 & 60.2 & 35.4 \\
    \midrule
    \textbf{Dynamic (Ours)}    & \textbf{47.8} & \textbf{66.8} & \textbf{40.8} \\
    \bottomrule
  \end{tabular}
\end{table}

\section{Conclusion}
We present PhyPrompt, a reinforcement learning-based framework that automatically enhances text-to-video prompts for physically plausible generation. Through a two-stage training approach combining physics-focused Chain-of-Thought supervision with dynamic reward-based reinforcement learning, PhyPrompt achieves substantial improvements in physical realism while maintaining semantic fidelity.
Our key empirical finding demonstrates that semantic adherence and physical commonsense are not inherently competing objectives when optimized through an appropriate curriculum. PhyPrompt-7B achieves 47.8\% semantic adherence and 66.8\% physical commonsense on the VideoPhy2 benchmark, yielding a 40.8\% joint success rate. Critically, this represents an 11 percentage point improvement in physical commonsense (from 55.8\% to 66.8\%) without sacrificing semantic quality. In fact, semantic adherence simultaneously improves by 4.4 percentage points (from 43.4\% to 47.8\%), resulting in an 8.6 percentage point gain in joint performance over baseline prompts.
This synergistic improvement stems from our dynamic curriculum mechanism, which discovers prompt-space regions unreachable through single-objective optimization. By establishing semantic scaffolding early in training before refining with physical constraints, the curriculum enables both objectives to mutually reinforce rather than trade off against each other. Our ablation studies confirm this effect: the dynamic curriculum outperforms not only static balanced weighting but also specialized single-objective training on each objective's own metric. Beyond performance gains, PhyPrompt demonstrates genuine zero-shot transfer across architecturally diverse generators (Lavie, VideoCrafter2, CogVideoX), achieving consistent improvements without per-generator fine-tuning. This model-agnostic design makes PhyPrompt a practical, deployable solution for physics-aware video generation.
Our results carry broader implications for multi-objective optimization in generative AI. The success of domain-specialized training with direct task feedback over raw parameter scaling suggests that careful curriculum design and targeted optimization can be more effective than simply increasing model size. As text-to-video systems continue advancing, prompt refinement approaches like PhyPrompt offer a parameter-efficient path toward physically grounded video synthesis suitable for robotics, scientific visualization, and educational applications where physical realism is essential.


{
    \small
    \bibliographystyle{ieeenat_fullname}
    \bibliography{main}
}

\clearpage
\setcounter{page}{1}
\maketitlesupplementary


\section{Example of Curated Data}
\cref{tab:cot-example} illustrates how we guide the model to inject physical reasoning: starting with a concise scenario description, we enforce the relevant law, clarify the expected visual outcome, and provide a step‐by‐step rationale that links microgravity and surface tension to the observed behavior of the liquid. This structured example demonstrates the depth and clarity of our CoT annotations.
\begin{table*}[h]
  \caption{Example from our curated physics‐focused Chain‐of‐Thought dataset. The original caption sets the scene in microgravity, the annotated law (‘Lack of gravity’) defines the physical constraint, the explicit caption specifies the outcome (spherical droplets floating), and the chain of thought.}
  \centering
  \begin{tabular}{@{}p{3cm}p{10cm}@{}}
    \toprule
    \textbf{Caption} & A can of soda is slowly poured out in the space station, releasing the liquid into the surrounding area. \\
    \midrule
    \textbf{Physical Law} & Lack of gravity. \\
    \midrule
    \textbf{Explicit Caption} & A can of soda is slowly poured out in the space station, releasing the liquid into the surrounding area. The liquid forms spherical droplets and floats freely in the air. \\
    \midrule
    \textbf{Chain of Thought} & In microgravity (or “zero-g” conditions), there is no dominant downward force pulling the liquid. Instead of flowing downward as it would on Earth, the soda drifts away once it exits the can. Surface tension becomes the main force acting on the soda droplets, pulling the liquid into the smallest possible shape, a sphere. With no significant gravitational force to break them apart, these droplets remain spherical and continue floating until they contact another surface or droplet. Hence, in the absence of gravity, soda poured in a spacecraft forms and floats as spherical droplets. \\
    \bottomrule
  \end{tabular}
  \label{tab:cot-example}
\end{table*}

\section{Video Generator Settings}
The evaluated teams converged on a similar fast-preview configuration, typically employing around 50 diffusion inference steps with classifier-free guidance to balance efficiency and visual quality. Specifically, CogVideoX-5B/2B utilize the DPM-Solver scheduler (CogVideoXDPMScheduler) configured with classifier-free guidance set to 6, producing 24-frame video clips (6 seconds at 4 fps) at a resolution of 480 × 720 pixels, all executed in bfloat16 precision \cite{yang2024cogvideox}. 
VideoCrafter2, by contrast, retains the classical DDIM sampler while increasing classifier-free guidance to 12, generating within the same 50-step inference budget video sequences of 24 frames (6 seconds at 4 fps) at 320 × 512 pixels in full fp32 precision \cite{wang2023lavie}. 
Similarly, LaVie leverages its base UNet architecture with classifier-free guidance set to 12, matching VideoCrafter2’s inference settings—50 DDIM steps to produce 24-frame sequences (6 seconds at 4 fps) at 320 × 512 resolution in fp32 precision \cite{chen2024videocrafter2}. 
All three diffusion-based models employ a cosine-based $\beta$ scheduling (Stable Diffusion’s standard schedule), utilize v-prediction parameterization, and explicitly follow a commonly adopted classifier-free guidance protocol with approximately 50 inference steps.

\section{Transferability of PhyPrompt to Different Video Generators}
\textbf{Results on Lavie.}
\cref{tab:enhanced_models} reports the performance of each prompt refinement strategy when generating videos with Lavie. Using the original prompts yields a baseline SA of 41.0\% and PC of 57.0\%, for a joint success rate of 29.2\%. Promptist underperforms the baseline, while simple instruction-tuned models (Qwen2.5) show modest gains in PC at the expense of SA. In contrast, PhyPrompt variants consistently improve both metrics – for example, PhyPrompt-7B raises PC to 63.2\% and joint success to 31.6\% – confirming its ability to inject physics-aware reasoning into prompts without degrading semantic alignment. GPT-4o and DeepSeek-V3 achieve near-baseline results, underscoring the unique advantage of our two-stage RL-trained rewriter on the Lavie generator.  
\begin{table}[ht]
  \centering
\caption{Semantic Adherence (SA), Physical Commonsense (PC), and joint SA\&PC scores for various prompt enhancement methods evaluated on the \textbf{Lavie} generator.}
  \label{tab:enhanced_models}
  \setlength{\tabcolsep}{3pt}
  \begin{tabular}{lccc}
    \toprule
    \textbf{Enhanced Model} & \textbf{SA (\%)} & \textbf{PC (\%)} & \textbf{PC \& SA (\%)} \\
    \midrule
    N/A (Original Prompt) & 41.0 & 57.0 & 29.2 \\
    Promptist             & 37.8 & 58.4 & 27.0 \\
    PhyT2V                & 41.2 & 60.8 & 30.0 \\
    \midrule
    Qwen2.5-1.5B          & 34.6 & 59.0 & 24.8 \\
    PhyPrompt-1.5B        & 40.2 & 61.2 & 30.6 \\
    Qwen2.5-3B            & 35.2 & 60.4 & 28.0 \\
    PhyPrompt-3B          & 41.4 & 60.8 & 30.2 \\
    Qwen2.5-7B            & 40.0 & 60.8 & 30.8 \\
    PhyPrompt-7B          & \textbf{41.8} & \textbf{63.2} & \textbf{31.6}\\
    \midrule
    GPT-4o                & 41.8 & 58.0 & 29.8 \\
    DeepSeek-V3           & 41.6 & 59.2 & 30.0 \\
    \bottomrule
  \end{tabular}
\end{table}

\textbf{Results on VideoCrafter2.}
\cref{tab:enhanced_vc2} reports results on VideoCrafter2. The original prompts achieve 45.6\% SA and 48.0\% PC for a 29.8\% joint score. Promptist underperforms the baseline, while Qwen2.5 variants trade off SA for higher PC (e.g., Qwen2.5-1.5B yields 35.0\% SA and 53.6\% PC). In contrast, PhyPrompt consistently boosts both metrics: the 1.5B variant raises joint success to 30.2\%, and the 7B variant attains the best scores (46.2\% SA, 66.4\% PC, 34.8\% joint). GPT-4o and DeepSeek-V3 deliver moderate PC improvements but fall short of PhyPrompt.
\begin{table}[ht]
  \centering
  \caption{Semantic Adherence (SA), Physical Commonsense (PC), and joint SA\&PC scores for various prompt enhancement methods evaluated on the \textbf{VideoCrafter2} generator.}
  \setlength{\tabcolsep}{3pt}
  \label{tab:enhanced_vc2}
  \begin{tabular}{lccc}
    \toprule
    \textbf{Enhanced Model} & \textbf{SA (\%)} & \textbf{PC (\%)} & \textbf{PC \& SA (\%)} \\
    \midrule
    N/A (Original Prompt) & 45.6 & 48.0 & 29.8 \\
    Promptist             & 43.8 & 51.4 & 27.2 \\
    PhyT2V                & 45.0 & 56.8 & 30.6 \\
    \midrule
    Qwen2.5-1.5B          & 35.0 & 53.6 & 27.8 \\
    PhyPrompt-1.5B        & 41.4 & 57.4 & 30.2 \\
    Qwen2.5-3B            & 40.0 & 55.8 & 28.0 \\
    PhyPrompt-3B          & 38.0 & 61.0 & 30.8 \\
    Qwen2.5-7B            & 45.8 & 58.0 & 31.2 \\
    PhyPrompt-7B          & \textbf{46.2} & \textbf{66.4} & \textbf{34.8} \\
    \midrule
    GPT-4o                & 38.4 & 60.0 & 30.0 \\
    DeepSeek-V3           & 40.0 & 60.4 & 32.8 \\
    \bottomrule
  \end{tabular}
\end{table}

\textbf{Results on CogVideoX-5B.}
\cref{tab:enhanced_cv5b} shows the performance of each prompt refinement strategy when driving the CogVideoX-5B generator. The original prompts achieve 50.2\% SA and 63.4\% PC (39.4\% joint). Promptist again underperforms the baseline, while Qwen2.5 models trade semantic fidelity for modest PC gains. 
In contrast, PhyPrompt variants largely restore SA and push PC higher: PhyPrompt-7B matches the baseline SA at 50.2\% and attains the best PC of 69.2\%, yielding a 42.0\% joint score. 
GPT-4o delivers the highest SA (52.6\%) but falls short of PhyPrompt on PC, and DeepSeek-V3 shows balanced but lower joint performance. 
These results confirm PhyPrompt’s ability to inject physics-aware reasoning into prompts for CogVideoX-5B without degrading semantic.

\begin{table}[ht]
  \centering
  \caption{Semantic Adherence (SA), Physical Commonsense (PC), and joint SA\&PC scores for various prompt enhancement methods evaluated on \textbf{CogVideoX-5B}.}
  \setlength{\tabcolsep}{3pt}
  \label{tab:enhanced_cv5b}
  \begin{tabular}{lccc}
    \toprule
    \textbf{Enhanced Model}   & \textbf{SA (\%)} & \textbf{PC (\%)} & \textbf{PC\&SA (\%)} \\
    \midrule
    N/A (Original Prompt)     & 50.2             & 63.4             & 39.4                   \\
    Promptist                 & 46.4             & 63.0             & 37.4                   \\
    PhyT2V                    & 50.6             & 65.0             & 40.8                   \\
    \midrule
    Qwen2.5-1.5B              & 39.2             & 66.4             & 31.0                   \\
    PhyPrompt-1.5B            & 47.4             & 66.6             & 39.0                   \\
    Qwen2.5-3B                & 45.2             & 64.6             & 35.8                   \\
    PhyPrompt-3B              & 46.4             & 68.4             & 40.2                   \\
    Qwen2.5-7B                & 47.8             & 63.8             & 38.2                   \\
    PhyPrompt-7B              & 50.2             & \textbf{69.2}             & \textbf{42.0}                  \\
    \midrule
    GPT-4o                    & \textbf{52.6}             & 66.4             & 41.8                   \\
    DeepSeek-V3               & 50.6             & 63.6             & 40.8                   \\
    \bottomrule
  \end{tabular}
\end{table}

\section{Training Settings}
\textbf{SFT Training Setup.}
We initialize from Qwen2.5-Instruct (1.5B/3B/7B)~\cite{Qwen2024}, which has been pretrained on diverse corpora and instruction-tuned for dialogue. 
To elicit explicit chain-of-thought reasoning, we insert token delimiters as follows:\texttt{<|im\_start|>think} and \texttt{<|im\_start|>answer}.
Each marker is surrounded by newlines to clearly separate “think” and “answer” stages.
We adopt the fine-tuning regimen of \cite{muennighoff2025simple}, optimizing the full model end-to-end for reasoning. 
Training runs for $5$ epochs with a per-GPU batch size of $16$. 
All computations use bfloat16 precision and a maximum sequence length of $4096$ tokens. 
We employ AdamW \cite{loshchilov2019decoupled} with $\beta_{1}=0.9$, $\beta_{2}=0.95$, weight decay $1\times10^{-4}$, and gradient clipping at $1.0$. 
The learning rate warms up linearly to $1\times10^{-5}$ over the first $5\%$ of steps and then decays to zero via a cosine schedule over the remaining steps.
We compute cross-entropy loss solely on reasoning traces and solutions, omitting question tokens. Fine-tuning is performed on a single node with $8\times\mathrm{A}100$ GPUs using Fully Sharded Data Parallel (FSDP) with CPU off-loading and gradient checkpointing; we save checkpoints every epoch and evaluate on a held-out validation set.

\textbf{GRPO Training Setup.} 
We fine-tune the policy LLM for a total of 250 steps using a learning rate of $1\times10^{-6}$, decayed linearly to zero. 
Training is conducted on a single node equipped with $8{\times}\mathrm{A}100$ GPUs, leveraging Fully-Sharded Data Parallel (FSDP) for efficient GPU memory usage, along with CPU off-loading and gradient checkpointing to further reduce memory footprint. 
During policy roll-outs, we utilize vLLM to generate \textbf{4} candidate responses per prompt, employing temperature $t=1.0$ and top-$p$ sampling with $p=1.0$ to ensure diversity. 
Input sequences are truncated to a maximum of 4096 tokens, while both generated responses and reasoning process are limited to at most 1024 tokens each. 
Policy optimization is performed using Generalized Advantage Estimation (GAE) with parameters $\lambda=1$ and $\gamma=1$, alongside a KL-regularization term controlled by coefficient $\beta=0.01$ and a maximum gradient norm of 20.
We set the per-GPU batch size to 1, resulting in an overall batch size of 8 across all GPUs, and checkpoints are saved every 100 training steps.

\section{Model Architecture}
\cref{tab:details} contrasts the architectures and prompt capacities of five video-generation and evaluation models. 
CogVideoX-5B and CogVideoX-2B \cite{wang2024cogvideodex} both pair a 3D causal VAE with a Transformer backbone (5B and 2B parameters, respectively), use a 4 096‐dimensional hidden state, and support long prompts of up to 226 tokens via RoPE or sinusoidal positional encodings. 
By comparison, LaVie \cite{wang2023lavie} employs an 860 M-parameter cascaded latent diffusion pipeline, VideoCrafter2 \cite{chen2024videocrafter2} uses a 1.2B-parameter UNet with temporal convolutions, and VideoPhy2 \cite{bansal2025videophy2} relies on a 7B-parameter Transformer enhanced with LoRA adapters; all three operate with a 768‐dimensional hidden state, a shorter 77‐token prompt window, and varied positional embeddings (RoPE, Fourier, or sinusoidal). 
This table highlights how model size, hidden dimension, position‐encoding strategy, and maximum prompt trade off across different approaches to physics-aware text-to-video synthesis and evaluation.

\begin{table*}[h]
\centering
\caption{Model details reported across the four papers.}
\label{tab:details}
\resizebox{\linewidth}{!}{
    \begin{tabular}{llllll}
    \toprule
    \textbf{Model} & \textbf{Core architecture} & \textbf{Size} & \textbf{Hidden Dim} & \textbf{Pos Enc} & \textbf{Max Prompt}\\
    \midrule
    CogVideoX-5B & 3D causal VAE $+$  Transformer & 5B & 4096 & RoPE & 226 \\
    CogVideoX-2B & 3D causal VAE $+$  Transformer & 2B & 4096 & sinusoidal & 226 \\
    Lavie & Cascaded Video Latent Diffusion  & 860M & 768 & RoPE & 77 \\
    VideoCrafter 2 &  UNet + temporal convolution & 1.2B & 768 & Fourier  & 77 \\
    VideoPhy2 & 7B Transformer with LoRA & 7B & 768 & sinusoidal & 77   \\
    \bottomrule
    \end{tabular}
}
\end{table*}

\end{document}